\pdfoutput=1

\documentclass[11pt]{article}

\usepackage[preprint]{latex/acl}

\usepackage{times}
\usepackage{latexsym}

\usepackage[T1]{fontenc}

\usepackage[utf8]{inputenc}

\usepackage{microtype}

\usepackage{inconsolata}

\usepackage{graphicx}
\usepackage{csquotes}

\usepackage{xcolor}
\usepackage{subcaption}  
\usepackage{booktabs}
\usepackage{multirow}
\usepackage{todonotes}
\usepackage{soul} 
\usepackage{marginnote}
\usepackage{url}

\usepackage[capitalize]{cleveref}
\Crefname{appendix}{App.}{Apps.}
\usepackage{listings}
\definecolor{TodoColor}{rgb}{1,0.7,0.6}
\definecolor{TodoColor2}{rgb}{0.7,0.7,0.9}
\definecolor{TodoColor3}{rgb}{0.5,0.8,0.5}

\def\TASKNAME{Speech-to-Abstract Generation}
\def\DATASETNAME{NUTSHELL}

\makeatletter\def\Hy@Warning#1{}\makeatother
\let\svthefootnote\thefootnote
\newcommand\blankfootnote[1]{%
  \let\thefootnote\relax\footnotetext{#1}%
  \let\thefootnote\svthefootnote%
}

\usepackage{footmisc}

\title{\DATASETNAME: A Dataset for Abstract Generation from Scientific Talks}

\author{
 \textbf{Maike Züfle\textsuperscript{1}},
 \textbf{Sara Papi\textsuperscript{2}},
 \textbf{Beatrice Savoldi\textsuperscript{2}},
 \textbf{Marco Gaido\textsuperscript{2}},
\\
 \textbf{Luisa Bentivogli\textsuperscript{2}},
 \textbf{Jan Niehues\textsuperscript{1}}
\\
\\
 \textsuperscript{1}Karlsruhe Institute of Technology,
 \textsuperscript{2}Fondazione Bruno Kessler
\\
 \small{\texttt{\{maike.zuefle,jan.niehues\}@kit.edu}, \texttt{\{spapi,bsavoldi,mgaido,bentivo\}@fbk.eu}
 }
}

\begin{document}
\maketitle

\begin{abstract}
Scientific communication is receiving increasing attention in natural language processing, especially to help researches access, summarize, and generate content.
One emerging application in this area is Speech-to-Abstract Generation (SAG), which aims to automatically generate abstracts from recorded scientific presentations. SAG enables researchers to efficiently engage with conference talks, but progress has been limited by a lack of large-scale datasets. To address this gap, we introduce \DATASETNAME{}, a novel multimodal dataset of *ACL conference talks paired with their corresponding abstracts. We establish strong baselines for SAG and evaluate the quality of generated abstracts using both automatic metrics and human judgments. Our results highlight the challenges of SAG and demonstrate the benefits of training on \DATASETNAME{}. By releasing \DATASETNAME{} under an open license (CC-BY 4.0), we aim to advance research in SAG and foster the development of improved models and evaluation methods.\footnote{\label{footnote_data_hf}\url{https://huggingface.co/datasets/maikezu/nutshell}}
\end{abstract}

\section{Introduction}
Abstracts are essential in scientific communication, allowing researchers
to quickly grasp the key contributions of a paper.
With the ever-growing number of publications, abstracts help researchers stay informed without reading full papers. Beyond their practical utility, abstracts also pose a significant challenge for natural language generation models: 
abstracts are a specialized form of summarization that not only condenses content but also promotes the work, often using domain-specific terminology and structured language.

Scientific summarization has been widely studied in natural language processing, including summarizing entire articles \citep{collins-etal-2017-supervised, mao-etal-2022-citesum, liu-etal-2024-sumsurvey}, particularly in the medical domain \citep{kedzie-etal-2018-content, cohan-etal-2018-discourse, gupta-etal-2021-sumpubmed}, generating abstracts from citations \citep{yasunaga-scisumm, Zanzotto_Bono_Vocca_Santilli_Croce_Gambosi_Basili_2020}, summarizing specific paper sections \citep{takeshita-etal-2024-aclsum}, and leveraging knowledge graphs for 
abstract generation \citep{koncel-kedziorski-etal-2019-text}.

\begin{figure}
    \centering
    \includegraphics[width=0.9\linewidth]{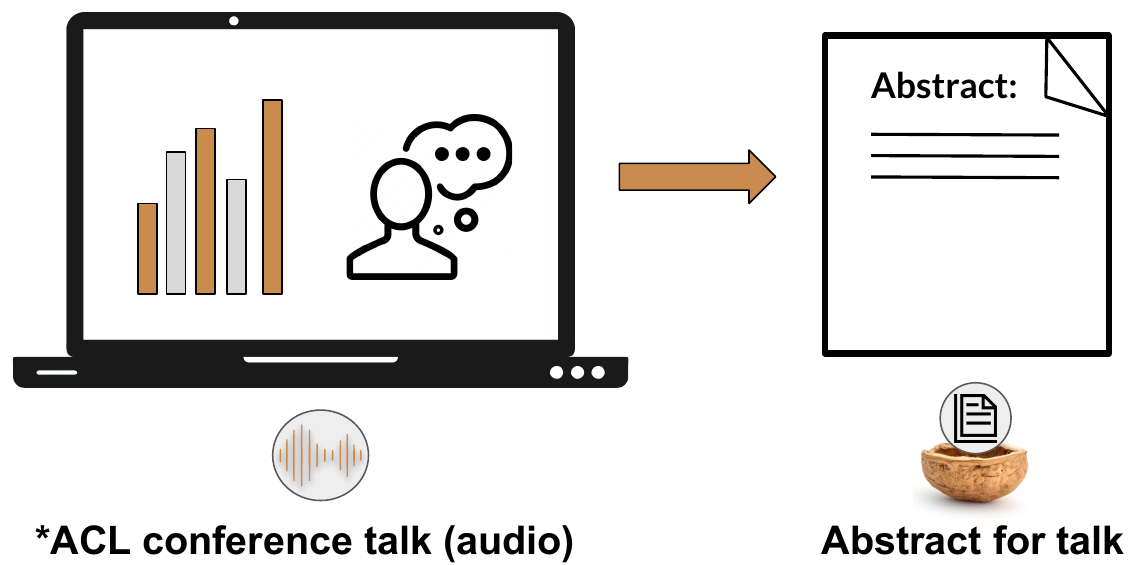}
    \caption{\DATASETNAME{}, a dataset for Speech-to-Abstract Generation (SAG) from scientific talks.}
    \label{fig:enter-label}
\end{figure}

With the growing availability of recorded conference talks, a new challenge emerges: generating abstracts from spoken content or \TASKNAME{} (SAG).  The abstracts offer researchers 
a quick way to assess relevant talks without watching entire recordings. Additionally, as conferences include more virtual content, automatically generated summaries enable efficient engagement with recorded talks \citep{murray-etal-2010-generating}.

While speech summarization has been explored in domains like news \citep{matsuura2024sentencewisespeechsummarizationtask}, YouTube videos \citep{sanabria2018how2largescaledatasetmultimodal}, and meeting minutes \citep{mccowan-ami, janin-icsi}, large-scale datasets for scientific talk abstract generation are lacking. 
Existing work \citep{lev-etal-2019-talksumm} aligns transcripts with the corresponding papers and extracts overlapping textual segments as summaries. However, these segments are drawn from the paper rather than the talk itself, failing to capture the distinct contributions, framing, and nuances conveyed in spoken presentations. Other studies have focused on summarizing TED Talks \citep{Koto-ted, DBLP:conf/asru/KanoODW21, vico-tedtalk-2022, shon-etal-2023-slue}, which target a broad audience and prioritize inspiration and engagement over technical content.

To bridge this gap, we introduce \DATASETNAME{} a new multimodal dataset for abstract generation from scientific talks. Built from recorded 
presentations of *ACL conferences, the dataset pairs abstracts with their corresponding spoken content and video, offering 
a valuable resource for future research. To validate the quality of the abstracts as concise and well-structured summaries of the talks -- i.e., capturing the essence of the presentations \textit{in a nutshell} -- we performed 
a human assessment, which confirmed
their effectiveness and suitability for the SAG task.

To establish baselines for SAG using our dataset, we evaluate three model types: (1) a cascaded model combining automatic speech recognition (ASR) with text-based summarization, (2) a state-of-the-art speech-language model (SpeechLLM) without fine-tuning, and (3) a SpeechLLM fine-tuned on our dataset. 

Our contributions are three-fold:
\begin{enumerate}
  \setlength{\itemsep}{1pt}
  \setlength{\parskip}{0pt}
  \setlength{\parsep}{0pt}
    \item  We introduce \DATASETNAME{}, a novel dataset for abstract generation from scientific talks comprising 1,172 hours, which is released under
    CC-BY 4.0 License on HuggingFace;\footref{footnote_data_hf}
    \item We provide baselines with different model types for comparison in future research, evaluated using both standard automatic metrics (e.g., ROUGE) and the emerging LLM-as-a-judge approach \citep{shen-etal-2023-large};
    \item We conduct human evaluations to assess the quality of 
        the abstracts and  validate the suitability of automatic metrics for the SAG task.

\end{enumerate}
\begin{table*}[!ht]
\small
    \centering
    \begin{tabular}{llcccccc}
    \toprule
     &  conferences & year & \# examples  & total audio & average audio & average words \\
     & & & &h & min & per abstract \\
     \midrule
        train  & ACL,NAACL, EMNLP & 2017-2021 & 4000 & 808.3 & 12.1 $\pm$ 11.2 & 142.8 $\pm$ 36.1 \\
         dev & ACL & 2022 & 885 & 146.4  & 9.9  $\pm$ 3.6 & 141.9 $\pm$ 36.5\\ 
         test & EMNLP, NAACL & 2022 & 1431 & 217.1 & 9.1  $\pm$ 4.3 & 147.6 $\pm$ 37.4 \\
    \midrule
    total & ACL, NAACL, EMNLP & 2017-2022& 6316 & 1171.8 & 11.1 $\pm$ 9.9 & 143.7 $\pm$ 36.5  \\
    \bottomrule
    \end{tabular}
    \caption{Dataset statistics for NUTSHELL. The number of words is obtained by splitting the abstract at whitespaces.}
    \label{tab:data_statistics}
\end{table*}

\section{The NUTSHELL Dataset}
\label{sec:dataset}
In this section, we introduce the new \DATASETNAME{} resource. We chose 
to build our corpus upon the
the ACL Anthology\footnote{\label{footnote_acl}https://aclanthology.org} 
since it provides a rich collection of multimodal resources (talks and abstracts) and open-access licensing. Starting from 2017, a significant number of papers published in the main *ACL conferences (ACL, EMNLP, and NAACL) include a video of the presentation, all released under the Creative Commons Attribution 4.0 license. This makes *ACL an ideal resource for building a multimodal dataset 
for the SAG task.

In the following, we present a feasibility assessment of SAG through human evaluation (\S\ref{subsec:human_feasibility}). Then, we describe the collection process performed to create \DATASETNAME{}, together with the final dataset statistics (\S\ref{subsec:collection}).

\subsection{Are paper abstracts \enquote{good} talk summaries?}
\label{subsec:human_feasibility}
Before creating the corpus,
we establish the validity of our data by investigating 
whether abstracts represent a good summary of the associated talk. To this aim, we conduct a qualitative check on a data sample of 30 talk-abstract pairs from the ACL Anthology. 
We involve a total of 5 annotators, who are all domain experts and thus familiar with scientific material.\footnote{Annotators include the paper authors and their colleagues.}
To verify Inter-Annotator Agreement (IAA),
a double annotation by different experts was carried out on 15 pairs.

Since we are interested in understanding whether 
paper abstracts are informative enough to represent a good summary
of the talk,
we asked evaluators to annotate:
    (1) Whether the information in the abstract is \textbf{all} uttered by the presenter in the talk;
    (2) The span of information present in the abstract that was not contained in the talk, if any;
    (3) Whether 
    the abstract summarizes 
    all \textbf{important} information presented in the talk.
The human evaluation procedure, including the annotation template, is described in \cref{app:human_eval_good_abstracts}.

The results indicate that $70.0\%$ of the abstracts are considered good summaries by annotators as they contain important information about the talk. 
However, $63.3\%$ of the abstracts also contain  information not explicitly present in the talk itself.
To better understand this, we conducted a qualitative analysis of the annotated spans corresponding to this missing information. 
We found that these spans typically involved dataset names, model names, shared task references (e.g., evaluation campaigns), or URLs (e.g., link to the resource or model being released). 
Notably, these elements are often displayed on slides but not explicitly verbalized by presenters.\footnote{This issue could be  overcome by exploiting the videos, as this information is typically shown in the slides. While out of scope for SAG, \DATASETNAME{} includes the videos, making it a useful resource also for more complex multimodal tasks.}

Despite this issue, the evaluation of automatic models against the same ground truth abstract can be considered fair, as models are equally penalized by this category of missing information. Moreover, it is worth noting that establishing a single ground truth for summarization tasks is still an open challenge  \citep{zhang-etal-2024-benchmarking}, given the inherent variability in human-produced summaries.

Both, questions (1) and (3) have an inter-annotator agreement of $\kappa=0.466$, indicating moderate agreement \citep{IAA-agreement}, which can be regarded as  acceptable given the subjective nature of evaluating summaries. 
While criterion (3) naturally involves subjective judgments about information importance, the lower agreement on criterion (1) can also be attributed to borderline cases, where small phrasing differences were sometimes overlooked by individual annotators. Such subtleties led to occasional discrepancies in annotator decisions, but were manually reviewed.

In summary, the manual evaluation confirmed both the feasibility of the SAG tasks and, despite the noted challenges, the overall reliability and usefulness of our resource.

\subsection{Collection and Dataset
Statistics}
\label{subsec:collection}
We collected talks from 16 ACL Anthology events: 6 ACL, 6 EMNLP, and 4 NAACL, including workshops, shared tasks and industry tracks.
For each paper (both long and short format), we extracted the video and the associated abstract already available on the paper website. We exclude papers with invalid URLs, videos without audio, or abstracts missing from the paper page. Additional details on the data collection can be found in \cref{sec:app:dataset_details}.

Lastly, we split the dataset into training (years  2017 to 2021), dev (ACL 2022), and test (EMNLP/NAACL 2022).
These splits reflect a realistic evaluation setup, where models are trained on past data and tested on the most recent, unseen examples.
In total, the corpus contains 1,172 hours of audio content corresponding to 6,316 different presentations. Full statistics are reported in \cref{tab:data_statistics}.

\section{Analysis}

 To demonstrate the quality and usability of our corpus, as well as 
 provide baselines for future works, we 
 develop and evaluate four different models using
 both automatic metrics and
 human evaluation. 

 \subsection{Experimental Setting}\label{subsec:exp_setting}

 \subsubsection{Models}
To establish baselines for the SAG task, we analyze the performance of four models described as follows. Prompts, model, generation, and additional training details are provided in \cref{sec:app:baselines}.

\paragraph{Whisper + LLama3.1-8B-Instruct.}  A cascaded solution, where the audio is first transcribed with
\texttt{openai/whisper\--large\--v3} \citep{radford2022robustspeechrecognitionlargescale}, and then
\texttt{meta\--llama/\-Llama\--3.1\--8B\--Instruct} \citep{dubey2024llama3herdmodels} is prompted to generate the abstract from the generated transcript.
\begin{table*}[!ht]
    \centering
    \resizebox{\linewidth}{!}{%
    \begin{tabular}{lcccccccccc}
    \toprule
       Model & \multicolumn{1}{c}{RougeL} & \multicolumn{1}{c}{BERTScore} & \multicolumn{3}{c}{Llama3.1-7B-Instruct}   & Human  (on subset) \\

      &   F1 $\uparrow$ &   F1 $\uparrow$ & Score with Expl. $\uparrow$ & Plain Score $\uparrow$ & Avg. Rank $\downarrow$ &  Avg. Rank $\downarrow$\\
    \midrule
       Whisper + LLama3.1-8B-Instruct  &  22.14 & 86.62 & \textbf{77.84} & \textbf{82.47} & \textbf{ 1.24} &  \textbf{1.53} \\
       Qwen2-Audio-7B-Instruct &  15.02 & 84.65 & 45.57 &   36.81 & 3.43 & 2.87\\
       End2End Finetuned &  \textbf{23.89} & \textbf{86.66} & 68.78 &  73.53 & 1.98 & 1.6\phantom{0} \\
       End2End Zero-Shot & 16.08 & 84.13 & 45.97 & 39.90 & 3.35 & N/A\\
    \bottomrule
    \end{tabular}%
    }
    \caption{We report results on the \DATASETNAME{} test set for four models: a cascaded approach (Whisper+Llama-3.1-8B-Instruct), an existing SpeechLLM (Qwen2-Audio), and an end-to-end \texttt{HuBERT+\-QFormer+\-Llama3.1-\-8B-\-Instruct} model, either finetuned on our data (\textit{End2End Finetuned }) or trained on audio instruction-following data (\textit{End2End Zero-Shot}). Avg. Rank, assigned by an LLM judge or human annotators, reflects the mean ranking per model. }
    
    \label{tab:baselines}
\end{table*}

\paragraph{Qwen2-Audio-7B-Instruct.} The \texttt{Qwen/\-Qwen2\--Audio\--7B\--Instruct} \citep{chu2024qwen2audiotechnicalreport} model, an existing SpeechLLM\footnote{By \textit{SpeechLLM}, we refer to the combination of a speech encoder and an LLM through a learned modality adapter \citep{gaido-etal-2024-speech}.}, which is used out of the box without any fine-tuning.

\paragraph{End2End Zero-Shot.} A SpeechLLM composed of HuBERT \citep{hubert-2021} as speech encoder, \texttt{meta\--llama/\-Llama\--3.1\--8B\--Instruct} as LLM, and a QFormer \citep{Li2023BLIP2BL} as adapter. The SpeechLMM is built to handle long audio inputs (\cref{sec:app:baselines}) and obtained by training only the adapter in two steps: (a) contrastive pretraining \citep{züfle2024contrastivelearningtaskindependentspeechllmpretraining} to align the LLM representations for the speech and text modalities using MuST-C \citep{di-gangi-etal-2019-must} and Gigaspeech \citep{chen-2021-gigaspeech}, and (b) fine-tuning on instruction-following tasks, including ASR, speech translation, and spoken question answering using MuST-C and Spoken-SQuAD \citep{lee2018spoken}. Therefore, the model is not trained or fine-tuned on \DATASETNAME{} and operates in zero-shot for the SAG task.

\paragraph{End2End Finetuned.} A SpeechLLM trained using the same contrastive pretraining procedure as End2End Zero-Shot but subsequently fine-tuned on our \DATASETNAME{} dataset. 
This not only evaluates the direct impact of task-specific datasets on the SAG performance, but it also ensures the feasibility of the task and the suitability of the collected data.

\subsubsection{Evaluation}
\paragraph{Metrics.} 
We use standard
(text) summarization metrics: \textbf{ROUGE} \citep{lin-2004-rouge} -- a text similarity metric that has been widely adopted for LM evaluation \citep{grusky-2023-rogue} that focuses on n-gram overlap between the hypothesis and reference --, and \textbf{BERTScore} \citep{DBLP:conf/iclr/ZhangKWWA20} -- a neural-based metric that measures the pairwise similarity of contextualized token embeddings between the summary and its reference. 
Also, we rely on \textbf{LLM-as-a-judge} \citep{shen-etal-2023-large,zheng-llm-judge-2024} 
 where the LLM\footnote{We use \texttt{Llama-3.1-8B-Instruct} \citep{dubey2024llama3herdmodels} as the judge using the prompts reported in \cref{fig:llm_as_a_judge} in \cref{sec:app-llm-as-a-judge}.} is prompted to assign a score to each output, using the reference abstract as context (Score with Expl.). The score is
  based on four criteria: 
(1) relevance, (2) coherence, (3) conciseness, and (4) factual accuracy.\footnote{(1) \textit{Does the predicted abstract capture the main points of the gold abstract?}, (2) \textit{Is the predicted abstract logically organized and easy to follow?}, (3) \textit{Is the predicted abstract free from unnecessary details?}, (4) \textit{Are the claims in the predicted abstract consistent with the gold abstract?}}
 We also report results where the LLM judge provides a single score without explanations (Plain Score), as well as results where it ranks the given abstracts instead of scoring them individually (Avg. Rank).
 
 All these metrics have known limitations and no metric is conclusively best for evaluating the SAG task: both ROUGE and BERTScore are known to fail to fully capture the extent to which two summaries share information \citep{deutsch-roth-2021-understanding} while LLM-as-a-judge is sensitive to prompt complexity and the length of input \citep{thakur2024judgingjudgesevaluatingalignment} and struggle to distinguish similar candidates \citep{shen-etal-2023-large}. For this reason, we complement 
the automatic scores with
 human evaluation.

\paragraph{Human Evaluation.}
For the human evaluation, 
nine annotators -- all experts in the field -- were provided 
with the generated abstracts and the ground truth abstract. We use the same randomly sampled 30 test set examples as in \cref{subsec:human_feasibility} and validate their representativeness, which is discussed in \cref{app:human_eval_ranking_model_outputs}.
Each sample is evaluated by three annotators. 
They follow the same criteria as the LLM evaluation but rank models instead of assigning scores. 
Detailed instructions are in \cref{app:human_eval_ranking_model_outputs}. 
As the End2End Zero-Shot model performance was comparable to that of Qwen2-Audio -- also being a zero-shot model -- and given that Qwen2-Audio is an established SpeechLLM with a distinct architecture, we exclude the End2End Zero-Shot from this analysis.

\subsection{Results}
\paragraph{Automatic Evaluation.}

\cref{tab:baselines} presents the performance of our models on the \DATASETNAME{} test set. Among them, the cascaded model (Whisper + Llama3.1-8B-Instruct) achieves the highest scores across all LLM-based evaluation metrics. Instead, looking at both n-gram- and neural-based metrics, the End2End Finetuned model achieves the highest RougeL and BERTScore. In addition, Qwen2-Audio and our End2End Zero-Shot models
demonstrate similar performance across all automatic metrics, showing a noticeable gap compared to the 
other two models. These results
highlight the importance of 
our dataset for building high-performing end-to-end models, as the substantial gap between the cascaded and End2End Zero-Shot models is effectively bridged through fine-tuning on the \DATASETNAME{} dataset.

For a more granular analysis, \cref{tab:baselines_with_llama_eval} in \cref{sec:app-llm-as-a-judge} provides 
results for the LLM-based 
metrics.
Given that all models except Qwen2-Audio rely on Llama3.1-8B-Instruct, one might 
question whether the Llama-based judge could introduce bias in favor of these models. To address this, we perform additional evaluations using \texttt{Qwen/Qwen2-7B} \citep{yang2024qwen2technicalreport} as the judge (\cref{tab:baselines_with_qwen_eval} in \cref{sec:app-llm-as-a-judge}), which confirm the same ranking,  eliminating any concerns about evaluator bias.

\paragraph{Human Evaluation.}

As shown in Table \ref{tab:baselines}, the human evaluation results closely align with the LLM-based  judgments: the cascaded model 
ranks first, followed closely by the finetuned model while Qwen2-Audio ranks last. Notably, the 
gap between the first two models is small, whereas the difference between the second and third models is substantial
-- consistent with the LLM-based evaluation. This suggests that automatic metrics reliably capture both subtle and large performance differences between models. IAA, measured using pairwise rankings \citep{bojar-etal-2016-findings} reached $\kappa = 0.53$, which is acceptable
given the close ranking of the top two systems.  

\section{Conclusion}\label{sec:conclusion}
In this work, we introduce \DATASETNAME{}, a novel dataset for SAG from recorded *ACL conference talks. By releasing this dataset under an open license, we hope to foster further advancements in SAG research and encourage the development of more effective models and evaluation techniques. 
Future work could explore the integration of the video content provided in the corpus, offering an additional modality for enriching the generation process and further improving abstract quality.
\section{Limitations}\label{sec:limitaitons}
While the current study provides a new resource and offers valuable insights about the SAG task, two main limitations should be noted:

\begin{itemize} 
\item The analysis focused on the 
speech-to-text abstract generation task. However, our dataset also provides 
access to the corresponding videos, which were not utilized here. Future research could explore the integration of video content as an additional modality to enhance the generation process and improve the quality of the abstracts. 
\item The human evaluation was limited in scope, involving only a small set of models and samples. Future work could expand this evaluation to include more models and a larger number of samples to better assess the performance of different metrics and determine which is most effective in various contexts. 
\end{itemize}

\paragraph{Potential Risks} 
Generating automatic summaries for scientific talks carries the risk that automatic summaries may misrepresent key findings or lack scientific accuracy. However, we hope that by providing more high-quality training data, summarization models can be improved and lead to more reliable and accurate summaries.

\section*{Acknowledgments}
This work has received funding from the European Union’s Horizon research and innovation programme under grant agreement No 101135798, project Meetween (My Personal AI Mediator for Virtual MEETtings BetWEEN People). We gratefully acknowledge Poland’s high-performance Infrastructure PLGrid ACC Cyfronet AGH for providing computer facilities. Beatrice Savoldi and Marco Gaido are supported by the PNRR project FAIR - Future AI Research (PE00000013), under the NRRP MUR program funded by the NextGenerationEU. 

We thank Leonard Bärmann, Béni Egressy, Mia Fuß, Lukas Hilgert, and Danni Liu for their contributions to the data annotation process.

\bibliography{custom, trimmed_anthology}

\clearpage
\appendix
\begin{figure*}[ht]
\textbf{System Prompt for Score with Explanation:}
\small{\begin{lstlisting}[breaklines=true, breakindent=0pt]
You are an expert AI trained to evaluate scientific abstracts. Your task is to compare a predicted abstract with a gold standard (reference) abstract and provide a detailed evaluation based on the following criteria:\n\n
1. **Relevance**: Does the predicted abstract capture the main points of the gold abstract?\n
2. **Coherence**: Is the predicted abstract logically organized and easy to follow?\n
3. **Conciseness**: Is the predicted abstract free from unnecessary details?\n
4. **Factual Accuracy**: Are the claims in the predicted abstract consistent with the gold abstract?\n\n
For each criterion:\n
- Assign a **score** between 1 and 10 (1 = very poor, 10 = excellent).\n"
- Provide a **brief explanation** for the assigned score.\n\n"
Your output must be in the following JSON format:\n\n"
{\"relevance\": {\"score\": int, \"explanation\": \"string\"},
 \"coherence\": {\"score\": int, \"explanation\": \"string\"},
 \"conciseness\": {\"score\": int, \"explanation\": \"string\"},
 W\"factual_accuracy\": {\"score\": int, \"explanation\": \"string\"}}\n\n
\end{lstlisting}}
\normalsize
\textbf{Prompt for Score with Explanation:}
\small \begin{lstlisting}[breaklines=true, breakindent=0pt]
### Gold Abstract:\n<reference abstract>\n\n### Predicted Abstract:\n<predicted abstract>\n\nPlease evaluate the predicted abstract based on the criteria mentioned.
\end{lstlisting}
\normalsize
\textbf{System Prompt for Score without Explanation:}
\small{\begin{lstlisting}[breaklines=true, breakindent=0pt]
You are an expert AI trained to evaluate scientific abstracts. Your task is to compare a predicted abstract with a reference abstract. Evaluate how well the prediction aligns with the reference using a score from 0 (lowest) to 100 (highest). Your output must only be in the following JSON format: {\"prediction\": int}. Do not provide any explanation or additional text.
\end{lstlisting}}
\normalsize
\textbf{Prompt for Score without Explanation:}
\small \begin{lstlisting}[breaklines=true, breakindent=0pt]
### Reference Abstract:\n<reference abstract>\n\n### Predicted Abstract:\n<predicted abstract>\n\nPlease evaluate the predicted abstract with respect to the reference abstract and assign a score from 0 to 100.
\end{lstlisting}
\normalsize
\textbf{System Prompt for Ranking:}
\small{\begin{lstlisting}[breaklines=true, breakindent=0pt]
You are an expert AI trained to evaluate scientific abstracts. Your task is to rank four different abstracts based on a reference abstract. Your output must only be in the following format: <Model A, Model B, Model C, Model D> where the first model is the best model, and the last model the weakest. Do not provide any explanation or additional text.
\end{lstlisting}}
\normalsize
\textbf{Prompt for Ranking:}
\small \begin{lstlisting}[breaklines=true, breakindent=0pt]
### Reference Abstract:\n<reference abstract>\n\n
### Model A Predicted Abstract:\n<predicted abstract 1>\n\n
### Model B Predicted Abstract:\n<predicted abstract 2>\n\n
### Model C Predicted Abstract:\n<predicted abstract 3>\n\n
### Model D Predicted Abstract:\n<predicted abstract 4>\n\n
Please rank the four predicted abstracts.
\end{lstlisting}

    \caption{Prompts for LLM as a judge. We use the same prompt for both, Qwen2-7bInstruct and Llama 3.1 8B Instruct. <reference abstract> and <predicted abstract> are replaced with the actual abstracts. For ranking, we shuffle the predicted abstracts, so that the LLMs sees the abstracts of different models in a different order every time to avoid position bias.}
    \label{fig:llm_as_a_judge}
\end{figure*}
\begin{table*}[!ht]
    \centering
    \resizebox{\linewidth}{!}{%
    \begin{tabular}{lccccccccccc}
    \toprule
       Model & \multicolumn{7}{c}{Llama-3.1-8B-Instruct}  \\

      & Relevance $\uparrow$ & Coherence $\uparrow$ & Conciseness $\uparrow$ & Factual Accuracy $\uparrow$ & Avg. Score $\uparrow$ & Plain Score $\uparrow$ &  Avg. Rank $\downarrow$\\
    \midrule
       Whisper + LLama31-Instruct  &  \textbf{77.12} & \textbf{86.00} & \textbf{61.13} & \textbf{87.13} & \textbf{77.84} & \textbf{82.47} & \textbf{1.24}\\
       Qwen2-Audio & 37.21 & 52.52 & 45.91 & 46.63 & 45.57 & 36.81 & 3.43\\
       End2End Finetuned & 66.41 & 78.24 & 50.25 & 80.22 & 68.78 & 73.53 & 1.98 \\
       End2End Zero-Shot & 40.28 &   48.02&  37.69 &  57.89&  45.97 & 39.90 & 3.35\\
    \bottomrule
    \end{tabular}%
    }
    \caption{Results using  Llama-3.1-8B-Instruct as a judge. We report results on the \DATASETNAME{} test set for four models: a cascaded approach (\texttt{openai/whisper-large-v3} + \texttt{meta/Llama-3.1-8B-Instruct}), \texttt{Qwen/Qwen2-Audio-7B-Instruct}, and an end-to-end \texttt{HuBERT+QFormer+Llama3.1-7B-Instruct} model, either finetuned on our data (\textit{End2End Finetuned }) or trained on audio instruction-following data (\textit{End2End Zero-Shot}). Avg. Rank reflects the mean ranking per model.}
    \label{tab:baselines_with_llama_eval}
\end{table*}

\begin{table*}[!ht]
    \centering
    \resizebox{\linewidth}{!}{%
    \begin{tabular}{lccccccccccc}
    \toprule
       Model & \multicolumn{7}{c}{Qwen2-7bInstruct} \\

      & Relevance $\uparrow$ & Coherence $\uparrow$ & Conciseness $\uparrow$ & Factual Accuracy $\uparrow$ & Avg. Score $\uparrow$ & Plain Score $\uparrow$ &  Avg. Rank $\downarrow$\\
    \midrule
       Whisper + LLama31-Instruct  & \textbf{79.61} & \textbf{83.54} & 72.08 & \textbf{86.07} & \textbf{80.33} & \textbf{74.60} & \textbf{1.66} \\
       Qwen2-Audio & 56.99 & 75.35 & \textbf{75.91} & 59.28 & 66.88 & 49.55 & 3.18\\
       End2End Finetuned & 75.13 & 81.78 & 75.04 & 81.16 & 78.28 & 70.83 & 2.12 \\
       End2End Zero-Shot & 57.93 & 68.02& 69.34 & 66.65& 65.49& 53.61 & 3.04 \\
    \bottomrule
    \end{tabular}%
    }
    \caption{Results using Qwen2-7bInstruct as a judge. We report results on the \DATASETNAME{} test set for four models: a cascaded approach (\texttt{openai/whisper-large-v3} + \texttt{meta/Llama-3.1-8B-Instruct}), \texttt{Qwen/Qwen2-Audio-7B-Instruct}, and an end-to-end \texttt{HuBERT+QFormer+Llama3.1-7B-Instruct} model, either finetuned on our data (\textit{End2End Finetuned }) or trained on audio instruction-following data (\textit{End2End Zero-Shot}). Avg. Rank reflects the mean ranking per model.}
    \label{tab:baselines_with_qwen_eval}
\end{table*}

\section{Human Evaluation: Are abstracts good summaries of the talk? }\label{app:human_eval_good_abstracts}
\begin{figure*}
    \centering
    \includegraphics[width=1\linewidth]{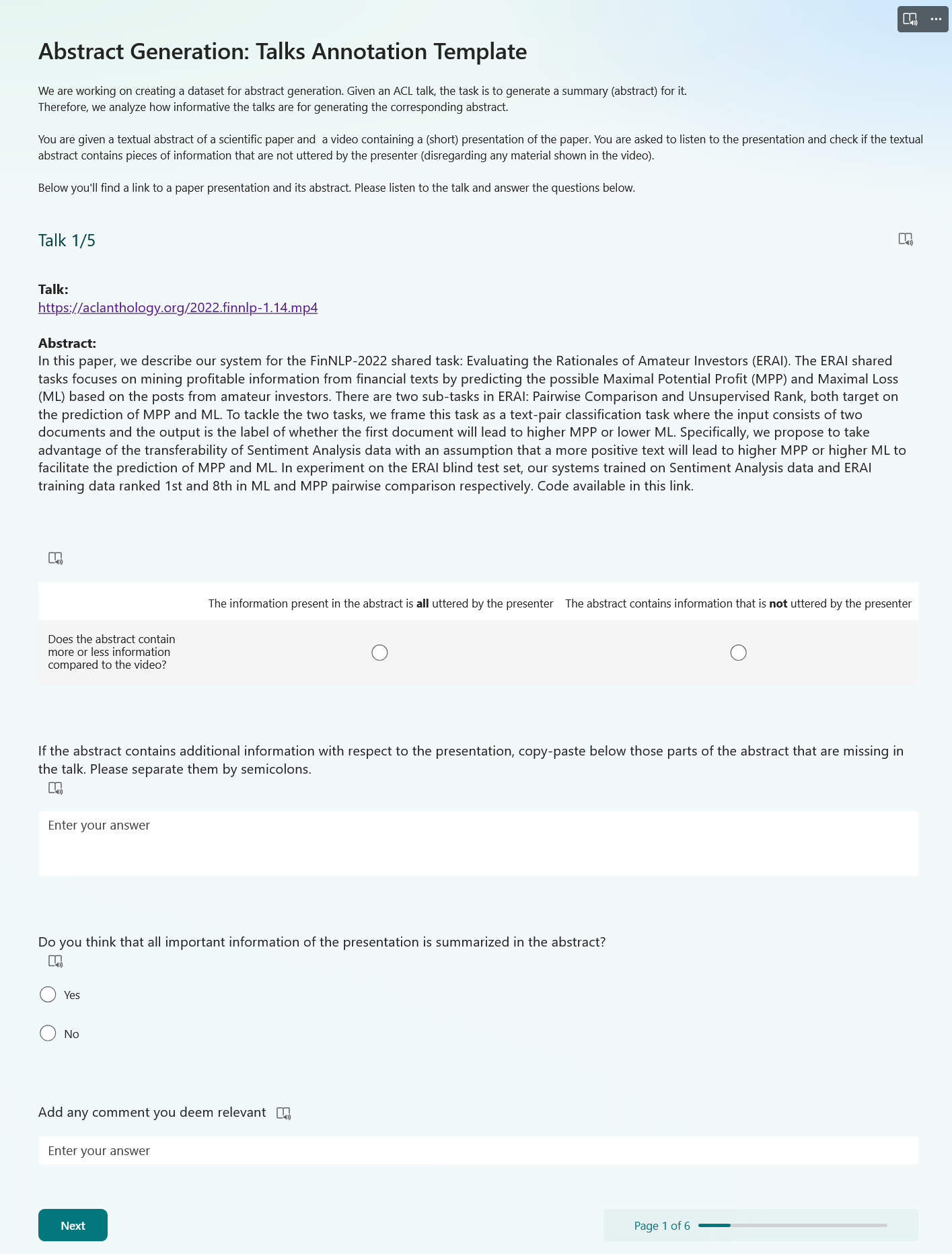}
    \caption{Instructions for annotators to evaluate whether the paper abstracts are good and informative abstracts for the ACL talks.}
    \label{fig:annoation_instructions_good_abstracts}
\end{figure*}

We aim to assess whether paper abstracts can serve as effective abstracts for *ACL talks. To this end, we conducted a human evaluation by randomly sampling 30 examples from our dataset. The annotation team consisted of five individuals (four women and one man), including the paper authors and their colleagues. All annotators were already familiar with the NLP domain, scientific presentation and writing, and the task itself. They are experts in Natural Language Processing, holding at least a master's degree in NLP or a related field, with some holding PhDs or professorial positions. Their ages ranged from 25 to 55.

The annotation guidelines were initially developed by the authors and subsequently refined in collaboration with the annotators to ensure a shared and well-defined set of evaluation criteria. Detailed instructions for the human annotators are provided in \cref{fig:annoation_instructions_good_abstracts}. 
The annotation template also included a comment section for uncertain cases, though no comments were submitted.

\section{Dataset Details}\label{sec:app:dataset_details}
We include all *ACL conferences from 2017 to 2022 in NUTSHELL, covering main conferences, Findings, industry tracks, and workshops. As not all conferences are held every year, the number of talks varies accordingly. \cref{tab:conference_talks} provides a detailed overview.
\begin{table}[ht]
\centering
\begin{tabular}{c|c|c|c}
\toprule
\textbf{Split} & \textbf{Conference} & \textbf{Year} & \textbf{Talks} \\
\midrule
\multirow{10}{*}{Train} 
  & \multirow{4}{*}{ACL} 
    & 2017 & 140 \\
  &   & 2018 & 185 \\
  &   & 2019 & 244 \\
  &   & 2021 & 849 \\
\cline{2-4}
  & \multirow{3}{*}{EMNLP}
    & 2017 & 93 \\
  &   & 2018 & 221 \\
  &   & 2021 & 1480 \\
\cline{2-4}
  & \multirow{3}{*}{NAACL}
    & 2018 & 120 \\
  &   & 2019 & 114 \\
  &   & 2021 & 554 \\
\hline
\multirow{1}{*}{Dev} 
  & ACL & 2022 & 885 \\
\hline
\multirow{2}{*}{Test} 
  & EMNLP & 2022 & 465 \\
  & NAACL & 2022 & 966 \\
\bottomrule
\end{tabular}
\caption{Number of talks per conferences in the NUTSHELL dataset.}
\label{tab:conference_talks}
\end{table}

\section{Baseline Details}\label{sec:app:baselines}
\paragraph{Generation Settings}
We evaluate four different models to establish baselines for abstract generation from spoken ACL talks. The evaluations were conducted on a single NVIDIA A100-SXM4-40GB GPU.

For all models, we use the default generation parameters and apply greedy search, following the usage instructions for \texttt{meta-llama/Llama-3.1-8B-Instruct}\footnote{\label{footnote_llama_url}\url{https://huggingface.co/meta-llama/Llama-3.1-8B-Instruct}} \citep{dubey2024llama3herdmodels}, \texttt{Qwen/Qwen2-Audio-7B-Instruct}\footnote{\label{footnote_qwen_url}\url{https://github.com/QwenLM/Qwen2-Audio}} \citep{chu2024qwen2audiotechnicalreport} and the contrastively pretrained models from \citet{züfle2024contrastivelearningtaskindependentspeechllmpretraining}\footnote{\label{footnote_llava_url}\url{https://github.com/MaikeZuefle/contr-pretraining}}.

\paragraph{Cascaded Model}
For the cascaded model, we segment the audio into 30-second chunks and transcribe them using \texttt{openai/whisper-large-v3} \citep{radford2022robustspeechrecognitionlargescale}. The transcribed chunks are then concatenated and processed by \texttt{meta-llama/Llama-3.1-8B-Instruct} \citep{dubey2024llama3herdmodels} to generate the abstract.  Inference took 5:40 hours on a single NVIDIA A100-SXM4-40GB GPU, including transcribing and summarizing.

Since the model’s outputs often included a title and category for the talk, we explicitly prompt it to generate only the abstract. This adjustment was not necessary for the other models.

We use the following prompt:

\lstset{basicstyle=\small} 
\noindent System Prompt:
\begin{lstlisting}[breaklines=true, breakindent=0pt]
A chat between a curious user and an artificial intelligence assistant. The assistant gives helpful, detailed, and polite answers to the user's questions.\n
\end{lstlisting}
Prompt:
\begin{lstlisting}[breaklines=true, breakindent=0pt]
Summarize the following talk to create an abstract for an ACL Paper, don't include the title or other information, only the abstract:\n<transcription>\n
\end{lstlisting}

\paragraph{Qwen2-Audio}
For \texttt{Qwen/\-Qwen2-\-Audio-\-7B-\-Instruct} \citep{chu2024qwen2audiotechnicalreport}, inference took 50 minutes on a single NVIDIA A100-SXM4-40GB GPU.
We use the system prompt as provided in the code documentation\footref{footnote_qwen_url}.

\lstset{basicstyle=\small} 
\noindent System Prompt:
\begin{lstlisting}[breaklines=true, breakindent=0pt]
You are a helpful assistant.
\end{lstlisting}
Prompt:
\begin{lstlisting}[breaklines=true, breakindent=0pt]
Summarize this talk to create an abstract for an ACL Paper:\n
\end{lstlisting}
\begin{table*}[!ht]
    \centering
    \resizebox{\linewidth}{!}{%
    \begin{tabular}{lccccccccc}
    \toprule
       Model & \multicolumn{1}{c}{RougeL} & \multicolumn{1}{c}{BERTScore} & \multicolumn{3}{c}{Llama3.1-7B-Instruct}   \\

      &   F1 $\uparrow$ &   F1 $\uparrow$ & Score with Expl. $\uparrow$ & Plain Score $\uparrow$ & Avg. Rank $\downarrow$ \\
    \midrule
       Whisper + LLama31-Instruct  &   23.26 &    86.81 &   \textbf{77.75} &  \textbf{84.30} & \textbf{1.23}  \\
       Qwen2-Audio &  16.26 &   84.94 & 48.42 &  39.50 & 3.47 \\
       End2End Finetuned &   \textbf{24.47} &  \textbf{86.71 }&   70.67 &   75.73 & 1.83\\
    \bottomrule
    \end{tabular}%
    }
    \caption{Baseline Results, the finetuned model is a HuBERT + Qformer + LLama31Instruct model on the subset used for human annotation (30 examples).}
    \label{tab:baselines_subset}
\end{table*}

\paragraph{Contrastively Pretained Models}
For the contrastively pretrained model, we follow \citet{züfle2024contrastivelearningtaskindependentspeechllmpretraining} and adopt their settings\footref{footnote_llava_url}, including training configurations, hyperparameters, and system prompts. The SpeechLLM consists of HuBERT \citep{hubert-2021} as speech encoder, \texttt{meta\--llama/\-Llama\--3.1\--8B\--Instruct} as LLM, and a QFormer \citep{Li2023BLIP2BL} as adapter. We choose HuBERT as an encoder in contrast to the bigger and more powerful \texttt{openai/\-whisper-\-large-\-v3} \citep{radford2022robustspeechrecognitionlargescale}, as it needs less memory and is therefore more suitable for the summarization task of longer audio. 
However, due to the extended duration of the audio inputs, we additionally introduce two modifications:
\begin{enumerate}
    \item     We segment the audio into one-minute chunks, encode each chunk using the encoder and then concatenate the encoded representations before passing them through the adapter and LLM backbone.
    \item     We use a batch size of 1 for fine-tuning with NUTSHELL.
\end{enumerate}

Despite these adjustments, we encountered memory limitations for audio files exceeding 35 minutes. In such cases, we truncate the audio to 35 minutes, which affects one example in the test set.

The training of the models was conducted on four NVIDIA A100-SXM4-40GB GPUs. The contrastive pretraining took 33 hours on four GPUS. Finetuning on ASR, speech translation, and spoken question answering data took 30 hours, finetuning on the NUTSHELL dataset took 2:10 hours.
Generating the outputs of the test set (on a single NVIDIA A100-SXM4-40GB GPU) took 2:35 hours.

\noindent System Prompt:
\begin{lstlisting}[breaklines=true, breakindent=0pt]
A chat between a curious user and an artificial intelligence assistant. The assistant gives helpful, detailed, and polite answers to the user's questions.\n
\end{lstlisting}
Prompt:
\begin{lstlisting}[breaklines=true, breakindent=0pt]
Summarize this talk to create an abstract for an ACL Paper:
\end{lstlisting}

\section{Evaluation Details}
We evaluate the results of our models using automatic metrics including ROUGE, BERTScore, and LLM-as-a-judge.
\subsection{ROUGE and BERT Score}
As automatic metrics, we use ROUGE\footnote{} \citep{lin-2004-rouge} and BERTScore  \citep{DBLP:conf/iclr/ZhangKWWA20}. Concretely, we compute ROUGE-L, which focuses on the longest common subsequence, with \texttt{DD/sacrerouge} \citep{deutsch-roth-2020-sacrerouge}, as recommended by \citet{grusky-2023-rogue} and for BERTScore, we use the \texttt{bertscore} implementation from HuggingFace\footnote{\url{https://huggingface.co/spaces/evaluate-metric/bertscore}} and report the F1-score.

\subsection{LLM as a judge}\label{sec:app-llm-as-a-judge}
To evaluate the model outputs, we also use an LLM \textit{as a judge}, specifically \texttt{meta\--llama/\-Llama\--3.1\--8B\--Instruct}  \citep{dubey2024llama3herdmodels}. The LLM assigns a score to each output using the reference abstract as context, based on four criteria: (1) relevance (\textit{Does the predicted abstract capture the main points of the gold abstract?}), (2) coherence (\textit{Is the predicted abstract logically organized and easy to follow?}), (3) conciseness (\textit{Is the predicted abstract free from unnecessary details?}), and (4) factual accuracy (\textit{Are the claims in the predicted abstract consistent with the gold abstract?}). Additionally, we report results where the LLM provides a single overall score without explanations and results where it ranks the given abstracts instead of scoring them individually. The prompts are given in \cref{fig:llm_as_a_judge}.
If the model fails to return a valid json dictionary, we instead take the first number after the score name in the output. 
We present the results for all four criteria, the average score, the score without explanations, and the ranking in \cref{tab:baselines_with_llama_eval}.  One potential concern is that this LLM might be biased, as all our models except Qwen2-Audio are based on Llama-3.1. However, we find this is not the case. When using \texttt{Qwen/Qwen2-7B} \citep{yang2024qwen2technicalreport} as the judge, we obtain the same ranking as with Llama. The results with Qwen-as-a-judge can be found in \cref{tab:baselines_with_qwen_eval}.

\section{Human Evaluation for Model Outputs}\label{app:human_eval_ranking_model_outputs}

We evaluate the models using ROUGE \citep{lin-2004-rouge}, BERTScore \citep{DBLP:conf/iclr/ZhangKWWA20}, and LLM-as-a-judge. However, it is known that automatic evaluation metrics can come with limitations. Namely, the first two metrics may not fully capture semantic overlap \citep{deutsch-roth-2021-understanding}, while LLM-as-a-judge is sensitive to prompt phrasing \citep{thakur2024judgingjudgesevaluatingalignment} and struggles to distinguish between closely similar candidates \citep{shen-etal-2023-large}. 
To validate the reliability of our automatic evaluation scores and better understand model behavior, we complement these metrics with a  human evaluation. This allows us also to verify the robustness of our findings.

\begin{figure*}
    \centering
    \includegraphics[width=1.0\linewidth]{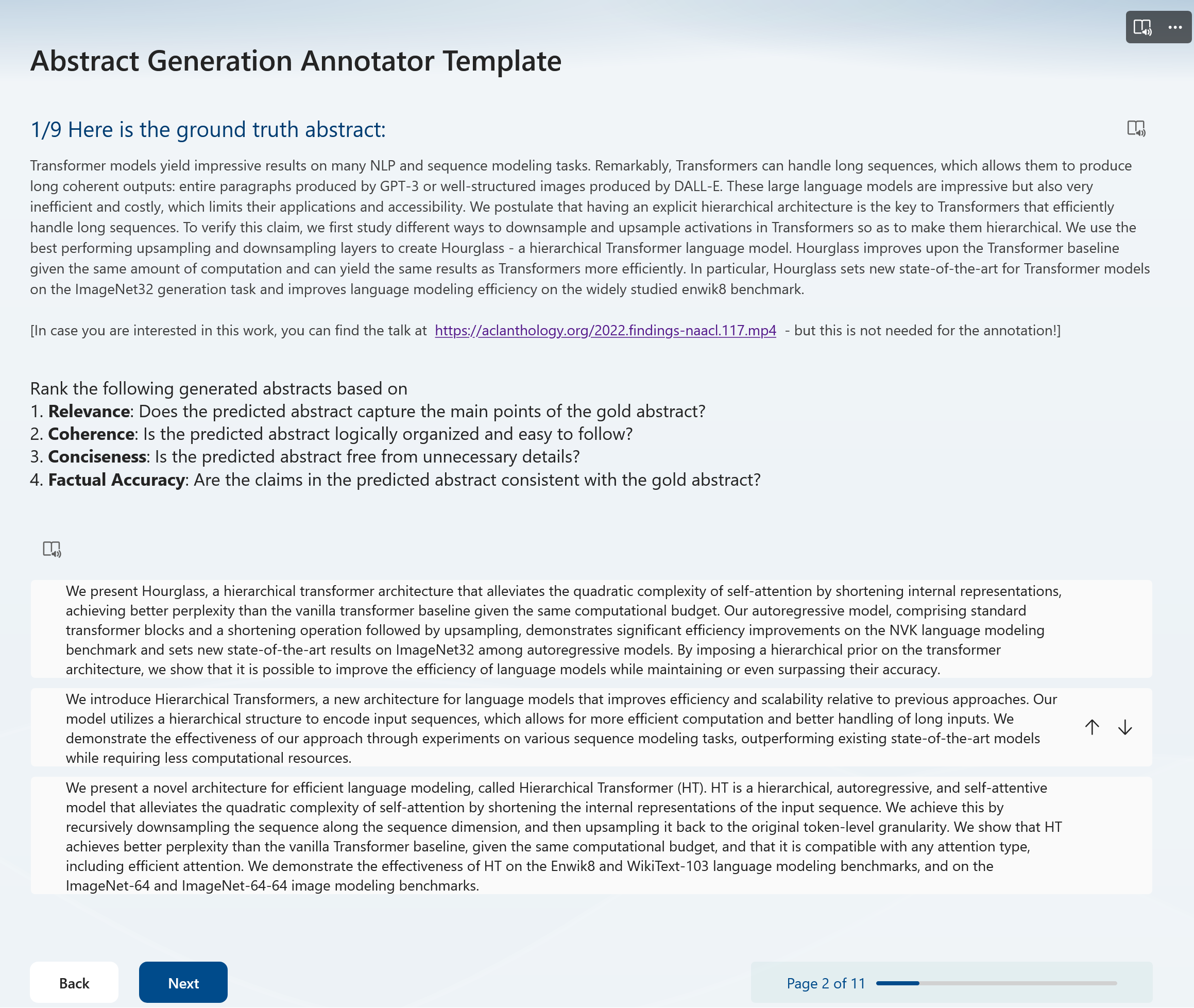}
    \caption{Instructions for human annotators for ranking model outputs.}
    \label{fig:annotation_instructions_ranking}
\end{figure*}

Specifically, we asked nine domain experts (four women and five men) to rank model outputs relative to the reference abstract, with each example annotated by three independent annotators. All annotators were already familiar with the NLP domain, scientific writing and presentation, and the task itself. They are experts in Natural Language Processing, holding at least a master's degree in NLP or a related field, with some holding PhDs or professorial positions. Their ages ranged from 25 to 55.
The annotation instructions are provided in \cref{fig:annotation_instructions_ranking}.

We conduct this human evaluation on a randomly selected subset of 30 test examples. We consider this subset representative, as the model rankings based on automatic metrics remain consistent with those on the full test set. The corresponding automatic scores for this subset are reported in \cref{tab:baselines_subset}. We want to include three diverse models in our human evaluation: a zero-shot model, a cascaded model, and a model finetuned on our dataset. Since we have two zero-shot models (Qwen2-Audio and our contrastively pretrained zero-shot model) that perform similarly, we decided to exclude one for efficiency in the human evaluation. We keep the Qwen2-Audio model as this is an already existing and widely used SpeechLLM.

\end{document}